\pdfoutput=1
\documentclass[11pt]{article}

\usepackage{acl}

\usepackage{listings}
\usepackage{times}
\usepackage{latexsym}
\usepackage[T1]{fontenc}
\usepackage[utf8]{inputenc}
\usepackage{microtype}
\usepackage{inconsolata}
\usepackage{graphicx}
\usepackage{booktabs}
\usepackage{multirow}
\usepackage{array}
\usepackage{colortbl}
\usepackage{makecell}
\usepackage{xcolor}
\usepackage{amsmath}
\usepackage{amsfonts}
\usepackage{kotex}
\usepackage[hang,flushmargin]{footmisc} 
\usepackage[most]{tcolorbox}
\usepackage{listings}

\title{DialBGM: A Benchmark for Background Music Recommendation from Everyday Multi-Turn Dialogues}

\author{Joonhyeok Shin$^{1}$, Jaehoon Kang$^{1}$, Yujun Lee$^{1}$, Hanna Lee$^{2}$, \\ \textbf{Yejin Lee$^{1}$, Kyuhong Shim$^{1}$} \\
$^{1}$Sungkyunkwan University, Republic of Korea \quad $^{2}$KAIST, Republic of Korea \\
\texttt{shinjh0729@skku.edu, khshim@skku.edu}
}
  
\begin{document}
\maketitle

\begin{abstract}
Background music (BGM) selection that fits natural human conversation is common in media production, yet automating it remains largely unaddressed. We introduce dialogue-conditioned BGM recommendation, the task of selecting non-intrusive music that fits a multi-turn conversation, even when the conversation contains no explicit music descriptors. We present DialBGM, a benchmark of 1,200 open-domain daily dialogues, each paired with four candidate music clips and human preference rankings under three suitability criteria: relevance, non-intrusiveness, and consistency. In our primary zero-shot candidate-ranking benchmark, no audio–language or multimodal LLM exceeds 35\% Hit@1, and inter-model ordinal agreement stays low (Kendall's $\tau_b < 0.4$). We then train a lightweight supervised score head on three multimodal LLM backbones; all improve over zero-shot, and two reach 0.49--0.52 Hit@1 under a dialogue-level split. These results suggest that DialBGM provides learnable supervision for dialogue-conditioned BGM preference, while track-disjoint generalization remains future work. DialBGM provides a benchmark for discourse-aware BGM recommendation, supporting both retrieval-based and generative model evaluation.
\end{abstract}

\section{Introduction}\label{sec:intro}

\begin{figure}[!t]
     \centering
      \includegraphics[width=1.0\linewidth]{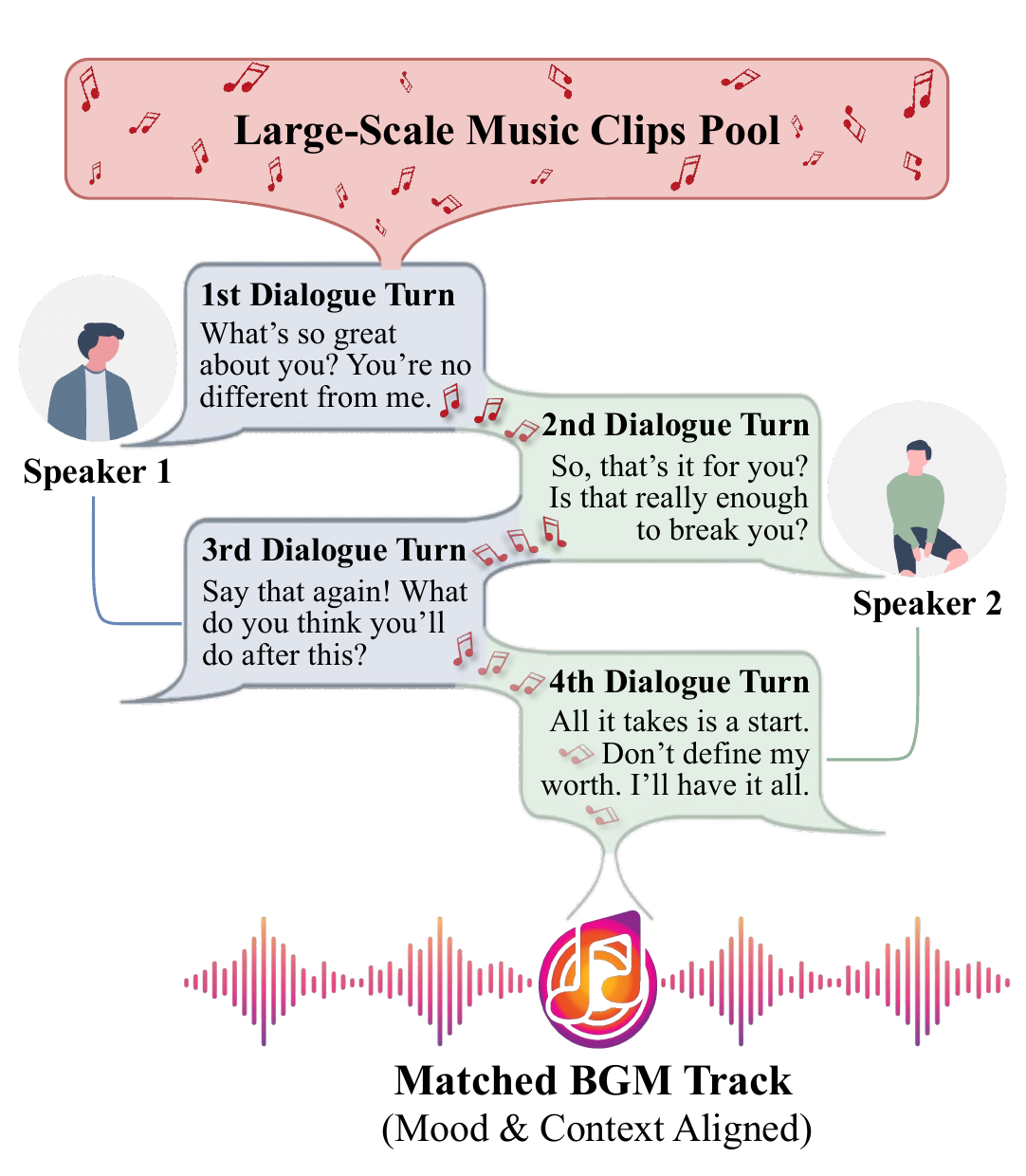}
      \caption{\textbf{Dialogue-conditioned BGM recommendation}. Given a multi-turn dialogue and a large-scale music clip database, the system uses the dialogue as a contextual filter to rank candidates and selects the one that best matches the dialogue as background music (BGM).}
      \label{fig:intro}
\end{figure}

 \begin{figure*}[t!]
     \centering
     \includegraphics[width=1.0\linewidth]{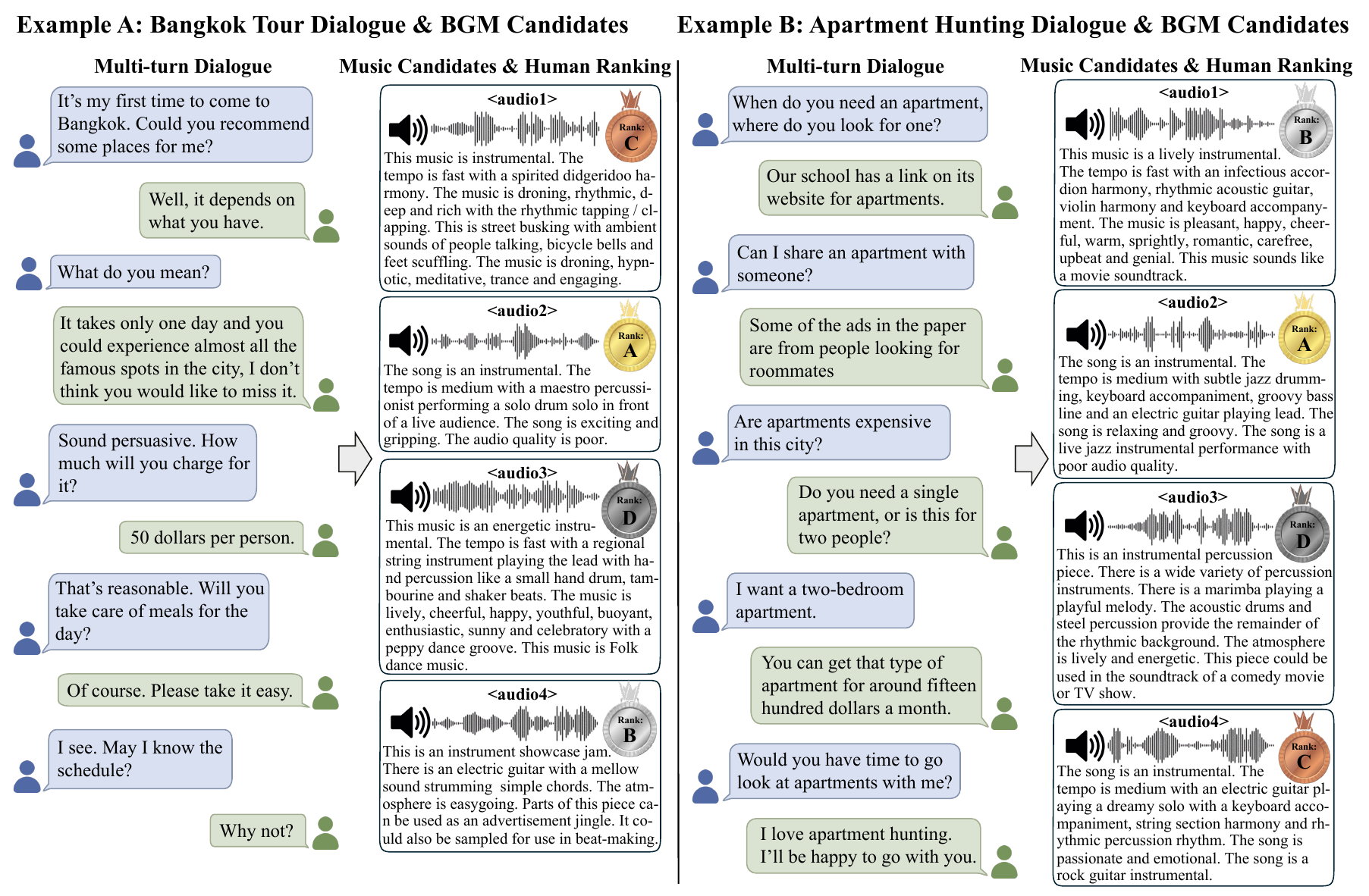}
     \caption{\textbf{DialBGM examples.} Each dataset instance consists of a multi-turn dialogue paired with four candidate background music (BGM) clips, along with human preference rankings indicating which clip best matches the conversational atmosphere. Each music clip is presented with its corresponding caption and human-annotated rank.}
     \label{fig:dataset}
 \end{figure*}

Background music (BGM) is widely used in movies, games, and interactive systems; it shapes atmosphere, guides attention, and strengthens emotional impact~\cite{ansani2020soundtracks}.
Despite its practical importance, automatically selecting suitable BGM from textual dialogues remains largely unaddressed.
In particular, to the best of our knowledge, there is no prior work that directly addresses dialogue-conditioned BGM recommendation, especially when the dialogue itself contains no explicit music descriptors.

Automation of BGM recommendation is intrinsically difficult because it requires bridging linguistic context and musical attributes.
Prior efforts in audio-text alignment have tackled music captioning and text-based music retrieval (e.g., CLAP~\cite{elizalde2023clap}, MuseChat~\cite{dong2024musechat}), but they often assume that the textual context contains sufficient clues like music descriptors (e.g., tags, artists, captions).
In contrast, selecting a suitable BGM for multi-turn daily dialogues is significantly more challenging.
Unlike general text-to-music retrieval, our input is conversational text that often contains no music descriptors, and our objective is background suitability (e.g., fit and non-intrusiveness), not simply semantic correspondence.
Unlike prior work on BGM selection for videos~\mbox{\cite{yi2021cross,mckee2023language}}, dialogue-conditioned BGM recommendation cannot rely on rich audiovisual signals or scene-level visual context. Instead, the model must infer latent affect and intent from conversational text alone.
The conversation may begin with tension and resolve warmly; it may convey sarcasm, embarrassment, or excitement through interaction patterns rather than explicit sentiment or musical words.
Notably, this application has practical significance for AI-driven media creation (e.g., adding soundtracks to dialogues in games, virtual reality, or storytelling applications) and for conversational agents that aim to enhance user experience with adaptive background ambience.

In this paper, we introduce a new task, \textbf{dialogue-conditioned BGM recommendation}, which requires selecting suitable background music for multi-turn conversations without explicit music descriptors.
Because the choice is conditioned on the dialogue alone, with no user profiles or listening histories, the task can also be understood as dialogue-conditioned BGM selection, distinct from personalized music recommendation.
To support this task, we present the \textbf{DialBGM} benchmark (see Figure~\ref{fig:intro}).
DialBGM pairs open-domain multi-turn dialogues with candidate music clips, utilizing human preference rankings to evaluate nuanced matching.
Specifically, each dialogue is paired with four candidate BGM clips, and the task is to rank the four music clips by how well they fit the dialogue (see Figure~\ref{fig:dataset}).
We systematically curate the candidate pool and employ human annotators to rank clips. 
The resulting dataset provides a consensus ranking for each dialogue’s quartet of music options, which supports retrieval/ranking-based evaluation and serves as a testbed for generative models.
DialBGM is designed primarily as a language-understanding evaluation resource: it extends DailyDialog~\cite{li2017dailydialog} with ranked BGM annotations that test whether a model can infer discourse-level affect, pragmatic intent, and conversational trajectory from a multi-turn dialogue.
The audio candidates provide the grounding space, but the central challenge is determining what the dialogue implies beyond isolated lexical cues.

Our experiments show that standard audio–text retrieval pipelines, such as summarizing the dialogue and retrieving music by caption similarity in a shared embedding space, remain inadequate for dialogue-conditioned BGM recommendation.
Across a wide range of open-source and proprietary models, no method achieves a Hit@1 score above 35\% in our primary zero-shot setting, highlighting a considerable gap relative to the aggregated human rankings.
These findings motivate a benchmark that explicitly targets complex dialogue understanding, rather than generic audio–text semantic alignment.

\vspace{0.1cm}
\noindent Our main contributions are as follows:
\begin{itemize}
    \item \textbf{Dataset and Construction Pipeline}: We present DialBGM, the first dataset that pairs multi-turn dialogues with background music, containing 1{,}200 daily dialogues, four music candidates per dialogue, and human preference rankings. We provide a reproducible four-stage construction pipeline (rule-based BGM filtering, LLM-based dialogue captioning, embedding-based candidate retrieval, human ranking) with explicit annotation criteria (relevance, non-intrusiveness, consistency), and we publicly release DialBGM.\footnote{\url{https://huggingface.co/datasets/joony97/DialBGM}}
    \vspace{-0.1cm}
    \item \textbf{Baseline and Analysis}: We evaluate a wide range of models, including embedding-based retrieval baselines (e.g., CLAP~\cite{elizalde2023clap}) and recent multimodal LLMs (e.g., Qwen2.5-Omni~\cite{xu2025qwen2}, Gemini 2.5~\cite{comanici2025gemini}, Music Flamingo~\cite{ghosh2025music}), and show that no system exceeds Hit@1 $=$ 0.35 with inter-model agreement below $\tau_b = 0.4$, establishing that DialBGM is hard.
    \vspace{-0.1cm}
    \item \textbf{Supervised Score Head}: We train a lightweight DoRA-adapted score head on three multimodal LLM backbones (Qwen2.5-Omni, Audio Flamingo 3, Phi-4-Multimodal) using a 1{,}000/200 dialogue-level split. All three improve over zero-shot, suggesting that DialBGM provides learnable supervision for dialogue-conditioned BGM ranking.
\end{itemize}
\section{Related Work}\label{sec:related}

\subsection{Audio-Language Models and Cross-Modal Retrieval}
Cross-modal retrieval between audio and text has advanced through contrastive learning approaches.
CLAP~\cite{elizalde2023clap} learns shared embeddings through contrastive learning, enabling zero-shot audio-text retrieval and classification.
LAION-CLAP~\cite{wu2023large} scales this approach with a large-scale dataset and feature fusion, achieving strong performance on audio-text retrieval benchmarks.
ParaCLAP~\cite{jing2024paraclap} extends this framework to paralinguistic attributes such as emotion and speaking style.
While these models capture acoustic characteristics, they are trained on short audio clips with brief captions, and their ability to align music with multi-turn dialogue context remains unexplored.

Beyond retrieval, large audio-language models (LALMs) have emerged for audio reasoning tasks.
Audio Flamingo~\cite{kong2024audio} supports few-shot learning and multi-turn dialogue, and its successor Audio Flamingo 3~\cite{ghosh2026audio} unifies speech, sound, and music understanding within a single open model, while Music Flamingo~\cite{ghosh2025music} specializes in music understanding via chain-of-thought reasoning.
Qwen2-Audio~\cite{chu2024qwen2} provides both voice chat and audio analysis modes, trained on diverse audio, including speech, music, and environmental sounds.
More recently, omni-modal models such as GPT-4o~\cite{hurst2024gpt}, Gemini 2.5~\cite{comanici2025gemini}, Qwen2.5-Omni~\cite{xu2025qwen2}, and Phi-4-Multimodal~\cite{abouelenin2025phi} integrate text, image, audio, and video understanding into unified architectures.
Despite recent progress, music understanding itself remains challenging, with models struggling to capture nuanced content~\cite{kang2025we}, and dedicated benchmarks such as MuChoMusic~\cite{weck2024muchomusic} show that audio--language models often over-rely on the language modality rather than the audio itself.

\subsection{Conversational Music Recommendation}
Traditional music recommendation relies on collaborative filtering or content-based approaches that focus on user listening history.
Early work, such as MusicRoBot~\cite{zhou2018musicrobot}, combined knowledge graphs with chatbots for context-aware music recommendation.
Talk the Walk~\cite{leszczynski2023talk} generates synthetic conversational data by performing biased random walks on playlist graphs, addressing data scarcity in conversational recommendation.
MuseChat~\cite{dong2024musechat} combines video understanding with dialogue-based refinement, explaining why specific tracks match visual content.
TalkPlay~\cite{doh2025talkplay} casts music recommendation as LLM token generation over multimodal music tokens, and TalkPlay-Tools~\cite{doh2025talkplaytools} extends it to an agentic setting in which the LLM invokes external tools such as SQL filters and dense retrieval.
More recently, MusiCRS~\cite{surana2026musicrs} introduced a benchmark for audio-centric conversational recommendation, linking music-seeking user conversations to candidate tracks.

These systems target explicit user preferences, playlist continuation, or music-seeking conversations.
In contrast, DialBGM focuses on implicit affective alignment: selecting BGM that matches the emotional trajectory of a conversation without explicit user requests, requiring models to interpret subtle mood cues rather than executing database queries.
Prior resources such as MusiCRS, TalkPlay, and MuChoMusic approach this space from music recommendation or audio modeling; DialBGM approaches it from discourse understanding.

\subsection{Datasets for Audio-Text Research}
High-quality paired datasets and benchmarks have driven progress in audio-language research~\cite{sakshi2024mmau, kumar2026mmau, lee2026beyond}.
For general audio captioning, AudioCaps~\cite{kim2019audiocaps} provides 46k human-annotated captions, while Clotho~\cite{drossos2020clotho} offers crowdsourced descriptions for 6k environmental sound clips.
WavCaps~\cite{mei2024wavcaps} further scales this effort to 400k clips using weakly labeled web data.

In the music domain, MusicCaps~\cite{agostinelli2023musiclm} provides 5.5k clips with professionally written captions.
LP-MusicCaps~\cite{doh2023lp} extends this to 2.2M samples using LLM-generated pseudo-captions.
MTG-Jamendo~\cite{bogdanov2019mtg} offers large-scale music tagging annotations, while Song Describer~\cite{manco2023song} provides free-form textual descriptions for music retrieval.

For emotion-aware dialogue, DailyDialog~\cite{li2017dailydialog} contains 13k conversations with emotion and dialogue act labels.
EmpatheticDialogues~\cite{rashkin2019towards} provides 25k dialogues grounded in emotional situations, and MELD~\cite{poria2019meld} offers multimodal emotion annotations from TV drama conversations.

Despite the availability of these resources, no existing dataset bridges multi-turn dialogue understanding with music selection.
DialBGM addresses this gap by pairing dialogues with human-ranked BGM candidates, enabling evaluation of affective alignment between conversational context and musical accompaniment.

\begin{figure*}[t!]
    \centering
    \includegraphics[width=1.0\linewidth]{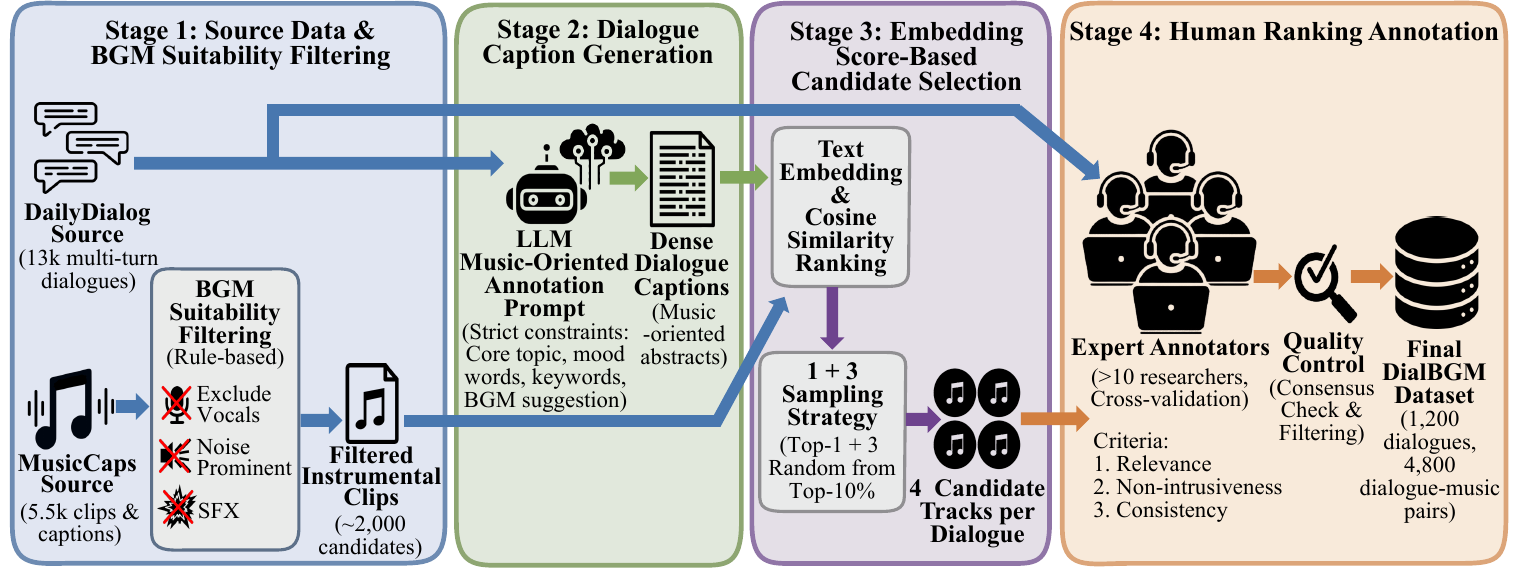}
    \caption{\textbf{Dataset construction.} The DialBGM dataset is constructed through a four-stage pipeline: (1) source data collection and rule-based BGM suitability filtering, (2) dialogue caption generation via an LLM, (3) embedding score-based candidate selection, and (4) expert human ranking annotation with quality control.}
    \label{fig:pipeline}
    \vspace{0.2cm}
\end{figure*}

\section{DialBGM Dataset}\label{sec:dataset}
DialBGM is constructed via a reproducible semi-automatic pipeline (Figure~\ref{fig:pipeline}): an automatic construction phase (Stages 1--3) and a human annotation phase (Stage 4).

\subsection{Task Definition and Evaluation Metrics}
\paragraph{Task Definition.}
Given a multi-turn dialogue $d$ and four music tracks $\mathcal{M} = \{m_1, m_2, m_3, m_4\}$ with varying semantic relevance, the model predicts a ranking permutation $R$. The reference is a consensus ranking $G$ from 1 (best) to 4 (worst), aggregated from 12 annotators (Section~\ref{sec:human_ranking}). Because BGM suitability is subjective, $G$ reflects what annotators agree is context-appropriate and non-intrusive; it is not a single objectively correct answer. This formulation tests whether a model identifies relative BGM preferences rather than retrieving a single matching item.

\paragraph{Evaluation Metrics.}
We adopt four metrics measuring distinct aspects of ranking alignment: Hit@1 (top-1 accuracy), MRR, nDCG (graded relevance with 3-2-1-0), and Kendall's $\tau_b$ (ordinal consistency). Formulas are listed in Table~\ref{tab:metrics}. Multimodal models frequently produce tied scores; we adopt tie-aware variants throughout: Hit@1 assigns $1/k$ for $k$-way ties, MRR uses expected reciprocal rank, nDCG averages over tie-permutations, and $\tau_b$ accounts for tied pairs ($T_x$, $T_y$).

\begin{table}[t!]
  \centering
  \caption{Metric formulas (idealized, no-tie). Tie-aware variants are described in Section~\ref{sec:dataset}.}
  \vspace{0.1cm}
  \label{tab:metrics}
  \renewcommand{\arraystretch}{1.2}
  \resizebox{\linewidth}{!}{
  \begin{tabular}{ll}
  \toprule
  \textbf{Metric} & \textbf{Formula} \\
  \midrule
  Hit@1 & $\mathbb{I}(\text{top}(R) = \text{top}(G))$ \\
  MRR & $\frac{1}{N} \sum_{i=1}^{N} \frac{1}{\text{rank}_i}$ \\
  nDCG & $\frac{\text{DCG}}{\text{IDCG}}, \quad \text{DCG} = \sum_{i=1}^{4} \frac{2^{rel_i}-1}{\log_2(i+1)}$ \\
  Kendall's $\tau_b$ & $\frac{P - Q}{\sqrt{(P + Q + T_x)(P + Q + T_y)}}$ \\
  \bottomrule
  \end{tabular}}
\end{table}

\begin{table}[t!]
  \centering
  \caption{Statistics of the DialBGM.}
  \vspace{0.1cm}
  \label{tab:dataset_stats}
  \resizebox{\linewidth}{!}{
  \setlength{\tabcolsep}{10pt}
  \begin{tabular}{l|c}
  \toprule
  \multicolumn{2}{c}{\textbf{Dialogue Statistics}} \\
  \midrule
  Total Dialogues & 1{,}200 \\
  Avg. Turns per Dialogue & 7.85 \\
  Avg. Words per Dialogue & 85.98 \\
  \midrule
  \multicolumn{2}{c}{\textbf{Audio Statistics}} \\
  \midrule
  Unique Audio Tracks & 1{,}020 \\
  Avg. Duration (sec) & 10.0 \\
  Avg. Instrument Families per Clip & 2.63 \\
  \bottomrule
  \end{tabular}}
\end{table}

\subsection{Automatic Dataset Construction Pipeline}

\paragraph{Source Data.} We leverage DailyDialog~\cite{li2017dailydialog} ($\sim$13,000 multi-turn dialogues with emotion/dialogue-act labels) and MusicCaps~\cite{agostinelli2023musiclm} (5,500 ten-second music clips with professional captions).

\paragraph{Stage 1: BGM Suitability Filtering.}
Not all MusicCaps clips are appropriate as background music. We apply a rule-based exclusion filter on the MusicCaps \texttt{aspect\_list} field, removing vocal-related terms (``vocal'', ``speech''), noise descriptors (``static'', ``hiss''), and SFX (``pop'', ``knock''). This yields $\sim$2,000 instrumental clips.

\paragraph{Stage 2: Dialogue Caption Generation.}
We next process the dialogue side. Multi-turn dialogues are verbose yet semantically sparse, so directly embedding raw text yields poor retrieval. We employ GPT-4o~\cite{hurst2024gpt} to distill each dialogue into a single-sentence music-oriented caption, conditioned on a BGM suggestion (style, instrument, tempo, energy). This Stage-2 caption is derived from the dialogue alone and serves as the retrieval query for candidate construction; Section~\ref{sec:experiment} reuses it as the one-line summary input, and it is not used to train the score head (Section~\ref{sec:supervised}). The full system prompt is in Appendix~\ref{app:dialogue_caption_prompt}.

\paragraph{Stage 3: Embedding Score-Based Candidate Selection.}
Using the generated captions, we treat BGM selection as text-to-text semantic retrieval over the 2,000 BGM-filtered captions, computing cosine similarity with OpenAI's \texttt{text-embedding-3-large}\footnote{\url{https://openai.com/index/new-embedding-models-and-api-updates/}}. We select four candidates per dialogue (six pairwise comparisons, manageable cognitive load); rather than top-4 by similarity (overly homogeneous), we adopt a ``1+3 Sampling Strategy'': top-1 plus three uniformly sampled from the top-10\%, intended to balance relevance with acoustic contrast.

\subsection{Human Ranking Annotation}
\label{sec:human_ranking}
Stage~3 retrieval captures surface semantics but cannot assess background-music suitability. We therefore collect human rankings on each retrieved quartet as the reference, which goes beyond textual similarity.

\paragraph{Annotation Protocol.}
We recruit 12 researchers with speech/audio experience as annotators. Each annotator independently views the complete dialogue with four BGM candidates, listens via a Gradio interface (Appendix~\ref{appendix:human}), and ranks 1 (best) to 4 (worst); the Stage-2 caption is not shown, so the consensus ranking is not mediated by it. We prefer ranking over absolute scoring because it reduces calibration issues and aggregates more reliably for subjective judgments.

\paragraph{Annotation Criteria.}
Three criteria: (1) \textit{Relevance} (mood/energy alignment with dialogue); (2) \textit{Non-intrusiveness} (background suitability; vocals/effects penalized); (3) \textit{Consistency} (stable atmosphere without abrupt changes).

\paragraph{Quality Control.}
We assess inter-annotator agreement via Kendall's $W$, with all 12 annotators per dialogue. Samples with $W < 0.25$ or in the bottom 10\% of agreement are excluded; this removes ${\sim}15\%$ of annotations, yielding 1{,}200 dialogues with mean $W = 0.79$, median $W = 0.90$, and 92.16\% achieving $W \geq 0.50$. Consensus rankings are aggregated via Borda count (Rank-1$\to$3, $\dots$, Rank-4$\to$0); ties are broken by the number of Rank-1 votes, then Rank-2 votes, and finally by candidate ID.

\subsection{Dataset Statistics}
Beyond the totals in Table~\ref{tab:dataset_stats}, dialogues span Daily Life \& Lifestyle (68\%), Work (19\%), Relationships (9\%), and others (4\%); emotional tones split into Neutral (41\%), Positive (36\%), Negative (14\%), and Mixed (9\%). The 1{,}020 unique audio tracks span Classical/Orchestral (15\%), Electronic (11\%), Jazz/Blues (10\%), and Rock/Metal (10\%), with Bass, Drums, and Guitar appearing most frequently.

\paragraph{Supervised split.}
For the supervised score-head experiments in Section~\ref{sec:supervised}, we use a dialogue-level $1{,}000/200$ train/test split. Unless otherwise specified, zero-shot benchmark results in Section~\ref{sec:experiment} are reported on all $1{,}200$ dialogues. Audio-track overlap across splits is analyzed in Section~\ref{sec:supervised} and discussed under Limitations.

\section{Benchmark Results}\label{sec:experiment}
\begin{figure}[!t]
  \centering
  \includegraphics[width=\linewidth]{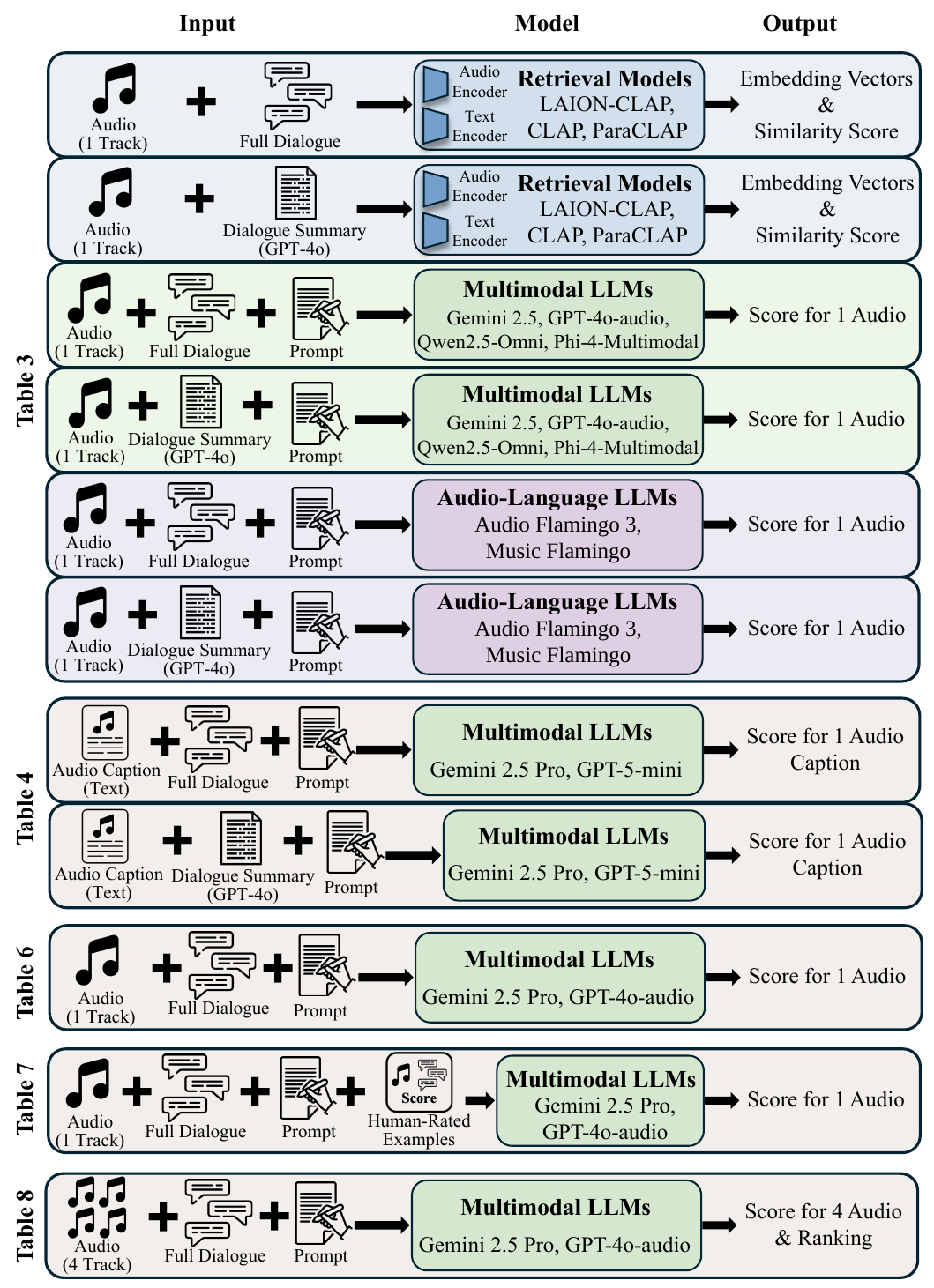
  }
  \caption{\textbf{Tested input-model-output configurations.} Overview of the experimental settings (Tables~\ref{tab:dialbgm_comparison}, \ref{tab:text_only_matching}, \ref{tab:app_no_cot}, \ref{tab:app_fewshot}, and~\ref{tab:app_joint_rerank}): input compositions (audio or audio caption; full dialogue or GPT-4o summary), model families (retrieval models, multimodal LLMs, audio-language LLMs), and output formats. Tables~\ref{tab:app_no_cot}--\ref{tab:app_joint_rerank} correspond to the advanced-prompting settings (without CoT, few-shot, and joint reranking).}
  \label{fig:exp_models}
\end{figure}

\begin{figure}[!t]
  \centering
  \includegraphics[width=\linewidth]{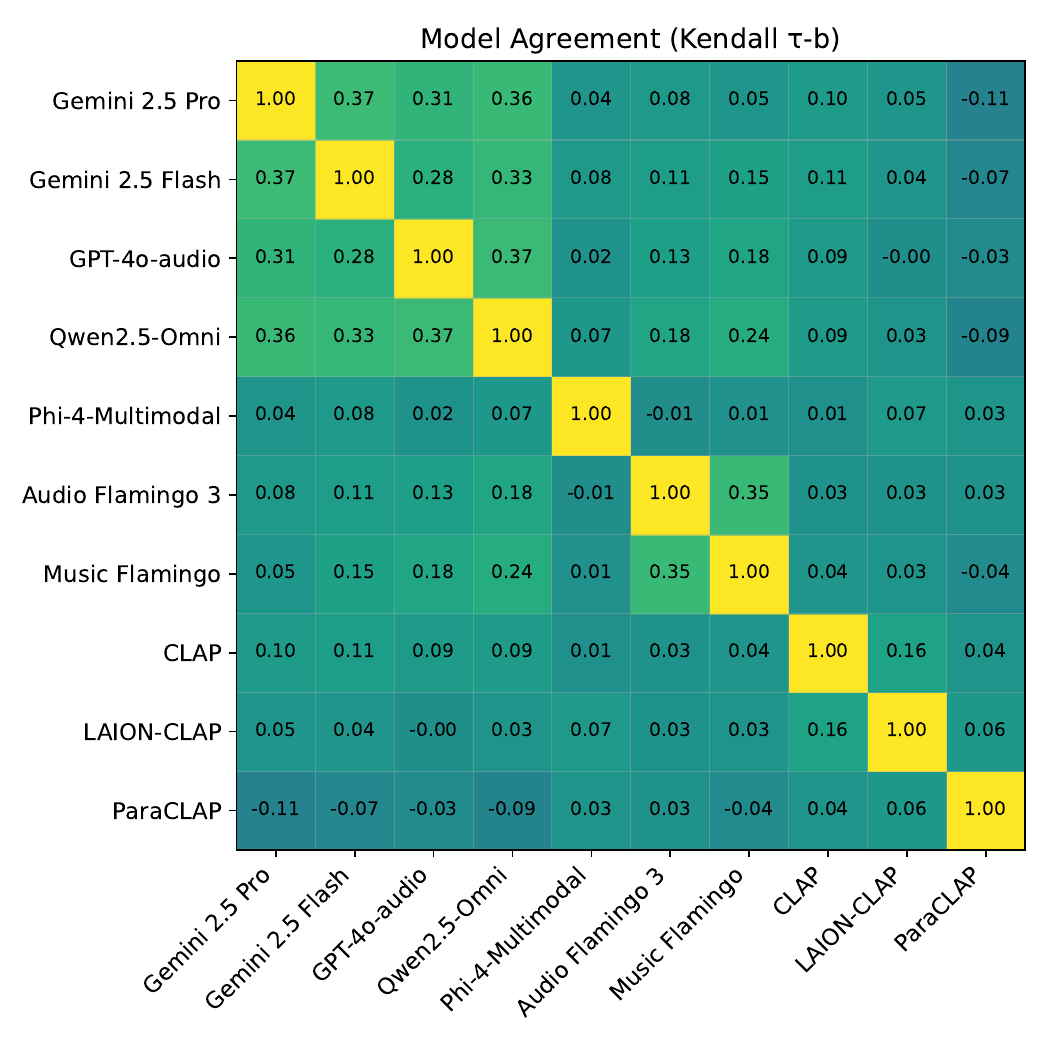}
  \caption{\textbf{Pairwise Kendall's $\tau_b$ across model rankings} on full-dialogue audio input. Intra-family clustering is visible; cross-family agreement is examined in Section~\ref{sec:inter_model_correlation}.}
  \label{fig:correlation_matrix}
\end{figure}

We evaluate contemporary multimodal LLMs, audio-language models, and audio-text retrieval models on all 1{,}200 DialBGM dialogues across two input configurations: an audio-text setting (full dialogues with audio candidates) and a text-only setting (dialogue captions with music captions). Figure~\ref{fig:exp_models} illustrates the configurations; pairwise model agreement is analyzed in Section~\ref{sec:inter_model_correlation} (Figure~\ref{fig:correlation_matrix}).

\subsection{Baseline Performance on DialBGM}

\begin{table*}[t]
    \caption{Performance of baseline models on the \textbf{DialBGM} Ranking task. $^\dagger$Accessed via the \texttt{gpt-audio} API. $^\ddagger$Candidates ranked by the cosine similarity between the Stage-2 dialogue caption and each music caption, i.e., the signal used to construct the candidate set; no audio is used.}
    \label{tab:dialbgm_comparison}
    \vspace{0.1cm}
    \setlength{\tabcolsep}{6pt}
    \centering
    \resizebox{1.0\linewidth}{!}{
    \begin{tabular}{llcccccccc}
        \toprule
        & & \multicolumn{4}{c}{\textbf{Full Dialogue}} & \multicolumn{4}{c}{\textbf{One-line Summary}} \\
        \cmidrule(lr){3-6} \cmidrule(lr){7-10}
        \textbf{Model Category} & \textbf{Model} & \textbf{Hit@1} & \textbf{MRR} & \textbf{nDCG} & \textbf{Kendall's $\tau_b$} & \textbf{Hit@1} & \textbf{MRR} & \textbf{nDCG} & \textbf{Kendall's $\tau_b$} \\
        \midrule
        \multirow{5}{*}{Multimodal LLMs}
            & Gemini 2.5 Pro    & 0.3233 & 0.5851 & 0.7904 & 0.1745 & 0.3474 & 0.5958 & 0.7965 & 0.1808 \\
            & Gemini 2.5 Flash  & 0.3293 & 0.5872 & 0.7905 & 0.1657 & 0.3228 & 0.5842 & 0.7912 & 0.1765 \\
            & GPT-4o-audio$^\dagger$ & 0.3051 & 0.5629 & 0.7822 & 0.1515 & 0.3192 & 0.5732 & 0.7856 & 0.1625 \\
            & Qwen2.5-Omni       & 0.3137 & 0.5742 & 0.7853 & 0.1739 & 0.3153 & 0.5740 & 0.7830 & 0.1674 \\
            & Phi-4-Multimodal   & 0.2545 & 0.5252 & 0.7524 & 0.0106 & 0.2493 & 0.5208 & 0.7525 & 0.0161 \\
        \midrule
        \multirow{2}{*}{\shortstack[l]{Audio-Language\\Models}}
            & Audio Flamingo 3   & 0.2651 & 0.5359 & 0.7657 & 0.0945 & 0.2799 & 0.5441 & 0.7694 & 0.1004 \\
            & Music Flamingo     & 0.2596 & 0.5306 & 0.7582 & 0.0598 & 0.2877 & 0.5518 & 0.7697 & 0.0965 \\
        \midrule
        \multirow{3}{*}{Retrieval Models}
            & CLAP               & 0.2392 & 0.5194 & 0.7529 & 0.0397 & 0.3033 & 0.5643 & 0.7795 & 0.1289 \\
            & LAION-CLAP         & 0.2708 & 0.5378 & 0.7612 & 0.0531 & 0.2933 & 0.5570 & 0.7748 & 0.1092 \\
            & ParaCLAP           & 0.2100 & 0.4924 & 0.7362 & -0.0264 & 0.2158 & 0.4957 & 0.7356 & -0.0442 \\
        \midrule
        Construction Baseline
            & Caption similarity$^\ddagger$ & \multicolumn{4}{c}{--} & 0.3383 & 0.5852 & 0.7844 & 0.1081 \\
        \bottomrule
    \end{tabular}
    }
\end{table*}

The evaluated models exhibit consistently low performance (Table~\ref{tab:dialbgm_comparison}): the best Hit@1 is $\approx 0.33$--$0.35$, MRR stays below 0.60, and Kendall's $\tau_b$ remains low ($-0.04$ to 0.18) despite relatively high nDCG (0.74--0.80); nDCG is less discriminative in this four-candidate setting because relevance scores are concentrated. Replacing the full dialogue with the one-line summary, which is the very query used to retrieve the candidates, yields only marginal changes. Table~\ref{tab:dialbgm_comparison} reports single-run scores. Multiple evaluation runs yield stable rankings, with Hit@1 standard deviations ranging from $\pm 0.002$ to $\pm 0.038$.

As a construction-time baseline, ranking each quartet by its Stage-3 caption similarity yields Hit@1 $= 0.338$ and $\tau_b = 0.108$ (last row of Table~\ref{tab:dialbgm_comparison}): its top-1 accuracy is comparable to the strongest zero-shot systems, but its ordinal agreement with the consensus is lower than that of the strongest multimodal LLMs ($\tau_b \approx 0.16$--$0.18$) and far below the supervised score heads in Section~\ref{sec:supervised} ($0.32$--$0.40$). The human rankings therefore capture dialogue--music suitability beyond the retrieval signal used to assemble the candidates.

One recurring failure mode illustrates the gap between model rankings and the consensus. When a speaker refuses a ``party'' because of academic failure, the models still rank ``spirited'' BGM highest: they latch onto a locally salient token (``party'') and miss the affective turn (the refusal and its cause) that sets the intended mood, consistent with surface-lexical matching rather than affect-grounded selection. Resolving such cases requires reading the dialogue as an emotional trajectory rather than a set of isolated cues, the competence our ranking task is designed to isolate and that caption-similarity retrieval cannot supply.

\subsection{Text-Only Does Not Close the Gap}
The lexical-overfitting pattern hints at a language-side bottleneck. To isolate it from audio perception, we conduct a text-only ablation with Gemini 2.5 Pro and GPT-5-mini~\cite{openai2025gpt5}, pairing music captions with full dialogues or summaries.

\begin{table}[t!]
    \caption{Text-only performance with full multi-turn dialogue or LLM-generated summary as input.}
    \vspace{0.1cm}
    \label{tab:text_only_matching}
    \centering
    \setlength{\tabcolsep}{5pt}
    \resizebox{\linewidth}{!}{%
    \begin{tabular}{lcccc}
        \toprule
        \textbf{Model} & \textbf{Hit@1} & \textbf{MRR} & \textbf{nDCG} & \textbf{$\tau_b$} \\
        \midrule
        Gemini 2.5 Pro \scriptsize{(full)}    & 0.3491 & 0.5978 & 0.8000 & 0.2003 \\
        GPT-5-mini \scriptsize{(full)}        & 0.3474 & 0.5945 & 0.7991 & 0.2016 \\
        \addlinespace[2pt]
        Gemini 2.5 Pro \scriptsize{(summary)} & 0.3321 & 0.5866 & 0.7968 & 0.1991 \\
        GPT-5-mini \scriptsize{(summary)}     & 0.3422 & 0.5954 & 0.8003 & 0.2106 \\
        \bottomrule
    \end{tabular}}
\end{table}

The text-only condition yields the highest scores yet remains low relative to the aggregated human rankings (Table~\ref{tab:text_only_matching}). This pattern is consistent with a limitation in affective reasoning rather than audio perception alone: models struggle to bridge dialogue narrative with the descriptive ``atmosphere'' of music.

\subsection{Limitations of Advanced Prompting}
If reasoning is the bottleneck, prompting should help. Yet across multimodal LLMs (GPT-4o-audio, Gemini 2.5), chain-of-thought~\cite{wei2022chain}, few-shot demonstrations with explicit guidelines~\cite{brown2020language}, and four-track joint reranking all fail to consistently improve over zero-shot; full results in Appendix~\ref{app:prompting_results}.

\subsection{Inter-Model Preference Alignment}
\label{sec:inter_model_correlation}

Even if individual models fall short, they might still share a notion of BGM suitability. We compute pairwise Kendall's $\tau_b$ correlations across rankings produced by all evaluated models under the full dialogue and audio input setting. Figure~\ref{fig:correlation_matrix} visualizes the resulting correlation matrix.

\paragraph{Architectural Clustering.}
Models with similar architectures exhibit higher agreement, forming distinct preference clusters. Multimodal LLMs (Gemini 2.5 Pro, Gemini 2.5 Flash, GPT-4o-audio, and Qwen2.5-Omni) show moderate positive correlations with one another ($\tau_b \approx 0.28$--$0.37$), which may reflect shared inductive biases from comparable pretraining objectives and model scale. In contrast, the Flamingo-based models (Audio Flamingo 3 and Music Flamingo) show strong intra-family agreement ($\tau_b \approx 0.35$) but weak alignment with the multimodal LLM cluster ($\tau_b \approx 0.05$--$0.24$).

\paragraph{Divergence of Retrieval Models.}
CLAP and its variants (LAION-CLAP, ParaCLAP) exhibit near-zero or negative correlation with the multimodal LLM cluster ($\tau_b \approx -0.11$--$0.11$), consistent with their reliance on static audio-text embedding similarity rather than affective reasoning.

\paragraph{Low Inter-Model Agreement.}
Despite the observed clustering, overall agreement remains low ($\tau_b < 0.4$ even for the most similar pairs). This absence of a shared preference structure across models is consistent with the difficulty of dialogue-conditioned BGM recommendation, but it should not be attributed to model failure alone: it may also reflect model-family biases, label ambiguity, multiple valid BGM choices for the same dialogue, or simple performance differences, which the present analysis cannot separate. Together with the gap to the annotated consensus, this motivates the supervised score head in Section~\ref{sec:supervised} rather than further zero-shot tuning.

\section{Supervised Score Head on DialBGM}\label{sec:supervised}

To probe whether DialBGM is learnable from training data alone, we train a lightweight supervised score head on three multimodal LLM backbones and evaluate on the 200-dialogue test split defined in Section~\ref{sec:dataset}. We compare each trained head against its zero-shot baseline, which is evaluated under the same generative-ranking protocol as the benchmark in Section~\ref{sec:experiment}.

\paragraph{Recipe.}
We evaluate Qwen2.5-Omni~\cite{xu2025qwen2}, Audio Flamingo 3~\cite{ghosh2026audio}, and Phi-4-Multimodal~\cite{abouelenin2025phi}. Qwen2.5-Omni and Audio Flamingo 3 use Qwen2.5-based language backbones and audio encoders initialized from Whisper-large-v3~\cite{radford2023robust}, with model-specific adaptation of both components. We also evaluate Phi-4-Multimodal, which pairs a Phi-4-Mini language backbone with a Conformer-based audio encoder~\cite{gulati2020conformer}. This lets us assess whether supervised ranking also improves performance for a backbone from a different language-model family. Each backbone produces a scalar score $s_i = \mathbf{w}^\top \mathbf{h}(d, a_i)$ via a linear head on the last-token hidden state; the four scores induce the predicted ranking. We adapt the LLM's attention and feed-forward (MLP) projections and the query, key, and value projections of the audio encoder with DoRA~\cite{liu2024dora}, an extension of LoRA~\cite{hu2022lora}, at rank $r=16$, $\alpha=32$, keeping the pretrained weights and the projector frozen. With reference permutation $\pi$ and graded reward $r_i \in \{3,2,1,0\}$, the head is trained with a pairwise Bradley--Terry loss~\cite{bradley1952rank} plus a listwise ListMLE term~\cite{xia2008listwise},
\begin{equation}
\mathcal{L} = -\tfrac{1}{6}\!\!\!\sum_{r_i > r_j}\!\!\! \log \sigma(s_i - s_j) - \lambda\! \sum_{t=1}^{3}\!\Big[ s_{\pi_t} - \log\!\!\!\sum_{j \in \mathcal{R}_t}\!\! e^{s_j} \Big],
\end{equation}
with $\mathcal{R}_t = \{\pi_t,\dots,\pi_4\}$ and $\lambda = 0.3$. We train for 3 epochs at learning rate $1{\times}10^{-4}$, an effective batch size of 8, cosine warmup 0.03, in bfloat16 on $2{\times}$A100 (40\,GB).

\begin{table}[t!]
\centering
\setlength{\tabcolsep}{6pt}
\caption{Supervised score head on the 200-dialogue test split. Zero-shot is evaluated under the same generative-ranking protocol as in Section~\ref{sec:experiment}.}
\resizebox{\linewidth}{!}{%
\begin{tabular}{llcc}
\toprule
\textbf{Backbone} & \textbf{Method} & \textbf{Hit@1} & \textbf{Kendall's $\tau_b$} \\
\midrule
\multirow{2}{*}{Qwen2.5-Omni}
 & Zero-shot       & 0.3188 & 0.2098 \\
 & + DoRA           & \textbf{0.5200} & \textbf{0.3979} \\
\midrule
\multirow{2}{*}{Audio Flamingo 3}
 & Zero-shot       & 0.2487 & $-0.0059$ \\
 & + DoRA           & \textbf{0.4875} & \textbf{0.3699} \\
\midrule
\multirow{2}{*}{Phi-4-Multimodal}
 & Zero-shot       & 0.2429 & 0.0029 \\
 & + DoRA           & \textbf{0.3950} & \textbf{0.3193} \\
\bottomrule
\end{tabular}}

\label{tab:supervised}
\end{table}

\paragraph{Results.}
All three backbones improve over their zero-shot baselines (Table~\ref{tab:supervised}). Qwen2.5-Omni reaches Hit@1 $=$ 0.5200 and Kendall's $\tau_b =$ 0.3979 ($+0.2012$ Hit@1, $+0.1881$ $\tau_b$), Audio Flamingo 3 reaches 0.4875 and 0.3699 ($+0.2388$, $+0.3758$), and Phi-4-Multimodal reaches 0.3950 and 0.3193 ($+0.1521$, $+0.3164$); all three exceed the best zero-shot Hit@1 of any model in our benchmark (Section~\ref{sec:experiment}, $\approx 0.33$--$0.35$). With DoRA restricted to the LLM, $\tau_b$ changes by less than 0.01 and Hit@1 by at most 0.05 for every backbone (single run), so the gain does not depend on adapting the audio encoder. The gain therefore holds across both audio encoders and LLM families; together, these results suggest that DialBGM provides learnable supervision for dialogue-conditioned BGM ranking.

\paragraph{What the score head learns.}
Two patterns in Table~\ref{tab:supervised} clarify what the supervision contributes. First, the largest relative gains are in ordinal consistency for the backbones whose zero-shot rankings are essentially uncorrelated with the consensus: Audio Flamingo 3 and Phi-4-Multimodal move from near-zero Kendall's $\tau_b$ to 0.37 and 0.32, respectively, indicating that supervision induces a consistent candidate ordering that the pretrained backbones did not expose zero-shot. Second, the two Qwen2.5-based backbones reach higher Hit@1 (0.49--0.52) than Phi-4-Multimodal (0.40), which is built on a different LLM family. Because Phi-4-Multimodal also exceeds the best zero-shot Hit@1 in our benchmark, the gain is not tied to a single audio encoder or LLM family, though our three backbones are too few to isolate which component drives the difference.

\paragraph{Controls for candidate priors and track reuse.}
The gain over zero-shot does not by itself show dialogue-conditioned reasoning: a head could also exploit dialogue-independent candidate priors or re-identify tracks that recur across splits. We quantify both on the 200-dialogue test split. First, a candidate-only prior that ranks each quartet by the candidates' mean human rank in the training split, without access to the dialogue, reaches Hit@1 $=$ 0.415 and $\tau_b =$ 0.279. This is a supervised prior rather than an input-side shortcut, and such a dialogue-independent component is expected by design, since non-intrusiveness and consistency are properties of the track. All three heads exceed this prior in $\tau_b$ (0.32--0.40), and Qwen2.5-Omni and Audio Flamingo 3 also exceed it in Hit@1. Second, because all candidates are drawn from a shared pool of 1{,}020 tracks, a fully track-disjoint split is not feasible, so we stratify the test split by track novelty. The Qwen2.5-Omni head scores 0.531 / 0.414 (Hit@1 / $\tau_b$) on the 145 dialogues whose candidates all appear in training and 0.491 / 0.354 on the 55 dialogues with at least one unseen candidate, where the prior scores 0.427 / 0.199. The supervised gain therefore cannot be explained solely by a candidate prior or by track re-identification, although a dialogue-level split cannot fully separate these effects.

\section{Conclusion and Future Work}\label{sec:conclusion}

\paragraph{Summary.}
DialBGM reframes BGM recommendation as a dialogue-understanding problem: systems must infer the latent affective trajectory of everyday conversation and judge which music can support it without becoming intrusive. We contribute the benchmark itself---1{,}200 dialogues with four ranked BGM candidates---together with a reproducible construction pipeline and a ranking-based evaluation protocol over Hit@1, MRR, nDCG, and Kendall's $\tau_b$.

\paragraph{Findings and Future Directions.}
Across audio-input, text-only, and prompting-based settings, current models remain weakly aligned with the human consensus rankings, and advanced prompting (chain-of-thought, few-shot) does not close the gap. That even the strongest text-only variant stays low suggests this capability does not emerge zero-shot from existing pretraining, but must instead be learned from task-specific supervision. As a first step, a lightweight supervised score head improves performance under a dialogue-level split, suggesting that DialBGM provides useful training signal while leaving track-disjoint generalization as an important next step. Future work will use DialBGM to train specialized audio-language adapters and to develop pretraining objectives for affective alignment between dialogue and music.

\section*{Limitations}
Despite the systematic construction and evaluation of the DialBGM benchmark, several limitations remain that future work should address.

\paragraph{Scale and Diversity of Data.} While DialBGM provides high-quality, human-verified rankings, the dataset size (1,200 dialogues) is relatively small compared to large-scale pretraining corpora. Consequently, the dataset is more suitable as an evaluation benchmark or a resource for fine-tuning, rather than for training LALMs from scratch. In addition, music candidates are restricted to the MusicCaps dataset, which implies that DialBGM may reflect biases present in MusicCaps.

\paragraph{Subjectivity of Annotations.} Although strict evaluation criteria are applied to achieve a high level of inter-annotator agreement, the suitability of BGM remains inherently subjective. For a given dialogue, a single or objectively "correct" music track rarely exists, and the rankings in DialBGM therefore reflect annotator consensus, which may not comprehensively capture the full spectrum of valid artistic interpretations. Our annotators are speech/audio researchers rather than professional music supervisors; incorporating professional supervisors, along with absolute-quality scoring and contrastive fitting/non-fitting labels, is left to future work.

\paragraph{Audio-Track Overlap Across Splits.} The 1{,}020 unique audio tracks are shared across the 1{,}200 dialogues, and our train/test split is defined at the dialogue level rather than the track level. Some tracks therefore appear in both splits, so the supervised results in Section~\ref{sec:supervised} should be read as dialogue-level rather than track-disjoint generalization.


\section*{Ethics Statements}
This work introduces DialBGM, a dataset for dialogue-conditioned BGM recommendation built on publicly available datasets (DailyDialog, MusicCaps); it contains no personally identifying or sensitive information.
Annotation was performed by 12 adult researchers in speech and audio who participated voluntarily; we did not recruit through a paid crowdsourcing platform and collected no demographic data, so the task involves only minimal-risk ranking judgments.
We release DialBGM annotations and metadata with source-specific license notices for DailyDialog and MusicCaps. We do not redistribute raw audio; downstream users should comply with the original dataset licenses and relevant copyright terms.
Overall, DialBGM is provided as a research resource that does not raise direct societal concerns and is constructed to facilitate progress in context-aware multimedia AI.

\section*{Acknowledgments}
This work was partly supported by the National Research Foundation of Korea (NRF) grant funded by the Korea government (MSIT) (RS-2026-25602529). This work was also partly supported by the Institute of Information \& Communications Technology Planning \& Evaluation (IITP) grants funded by the Korea government (MSIT) (IITP-2026-RS-2026-25615334, AI Star Fellowship Support Program; RS-2019-II190421, Artificial Intelligence Graduate School Program (Sungkyunkwan University)). This work was also partly supported by the Institute of Information \& Communications Technology Planning \& Evaluation (IITP, AI Computing Support Project for R\&D) grant funded by the Korea government (MSIT) (High-Performance Research AI Computing Infrastructure Support at the 2 PFLOPS Scale, 25\%).

\bibliography{custom}

\newpage
\appendix
\newtcolorbox{promptbox}[1][]{
  colback=gray!5,
  colframe=gray!40,
  boxrule=0.5pt,
  arc=2mm,
  left=1em, right=1em, top=1em, bottom=1em,
  fontupper=\small\ttfamily,
  title=\textbf{#1}
}

\section{Additional Experimental Results for Advanced Prompting}
\label{app:prompting_results}

\paragraph{Chain-of-Thought (CoT).}
Table~\ref{tab:app_no_cot} reports results obtained without CoT prompts under the same evaluation setting.
When compared to the CoT-based results in Table~\ref{tab:dialbgm_comparison}, performance differences remain minor across all metrics, indicating that explicit reasoning decomposition does not provide consistent gains.

\begin{table}[h!]
    \centering
    \caption{Results without Chain-of-Thought prompting.}
    \resizebox{\columnwidth}{!}{%
        \begin{tabular}{lcccc}
            \toprule
            \textbf{Model} & \textbf{Hit@1} & \textbf{MRR} & \textbf{nDCG} & \textbf{Kendall's $\tau_b$} \\
            \midrule
            Gemini 2.5 Pro & 0.3712 & 0.6193 & 0.8103 & 0.2544 \\
            GPT-4o-audio        & 0.3142 & 0.5695 & 0.7837 & 0.1578 \\
            \bottomrule
        \end{tabular}
    }
    \label{tab:app_no_cot}
\end{table}

\paragraph{Few-Shot Examples and Guidelines.}
Table~\ref{tab:app_fewshot} presents results using few-shot demonstrations with human-labeled scores and explicit guidelines (Relevance, Non-intrusiveness, Consistency).
Relative to the corresponding baseline in Table~\ref{tab:dialbgm_comparison}, the differences remain small, suggesting that this form of example-based instruction does not reliably improve performance.

\begin{table}[h!]
    \centering
    \caption{Results with few-shot demonstrations and explicit guidelines.}
    \resizebox{\columnwidth}{!}{%
        \begin{tabular}{lcccc}
            \toprule
            \textbf{Model} & \textbf{Hit@1} & \textbf{MRR} & \textbf{nDCG} & \textbf{Kendall's $\tau_b$} \\
            \midrule
            Gemini 2.5 Pro & 0.3443 & 0.5976 & 0.7941 & 0.1758 \\
            GPT-4o-audio        & 0.2949 & 0.5553 & 0.7769 & 0.1365 \\
            \bottomrule
        \end{tabular}
    }
    \label{tab:app_fewshot}
\end{table}

\paragraph{Joint Ranking (Four-Track Comparison).}
Table~\ref{tab:app_joint_rerank} reports results for joint reranking, where the model receives all four candidate clips simultaneously and outputs a single global ordering.
Consistent with Section~\ref{sec:experiment}, joint reranking does not yield reliable improvements across models.

\begin{table}[h!]
    \centering
    \caption{Joint reranking results where all four audio candidates are provided simultaneously.}
    \resizebox{\columnwidth}{!}{%
        \begin{tabular}{lcccc}
            \toprule
            \textbf{Model} & \textbf{Hit@1} & \textbf{MRR} & \textbf{nDCG} & \textbf{Kendall's $\tau_b$} \\
            \midrule
            Gemini 2.5 Pro & 0.3204 & 0.5848 & 0.7924 & 0.1783 \\
            GPT-4o-audio & 0.3058 & 0.5695 & 0.7849 & 0.1690 \\
            \bottomrule
        \end{tabular}
    }
    \label{tab:app_joint_rerank}
\end{table}

\section{Data Collection Interface}~\label{appendix:human}
\label{sec:appendix_interface}
To facilitate efficient data collection, we implement a lightweight web-based annotation tool using Gradio. For each dialogue-music candidate set, the interface presents (i) the full multi-turn dialogue context and (ii) four candidate audio tracks. Annotators listen to the candidates and assign a complete ranking from 1 (best) to 4 (worst) according to the task definition (selecting background music that best supports the dialogue).

\begin{figure*}[ht!]
    \centering
    \includegraphics[width=0.9\linewidth]{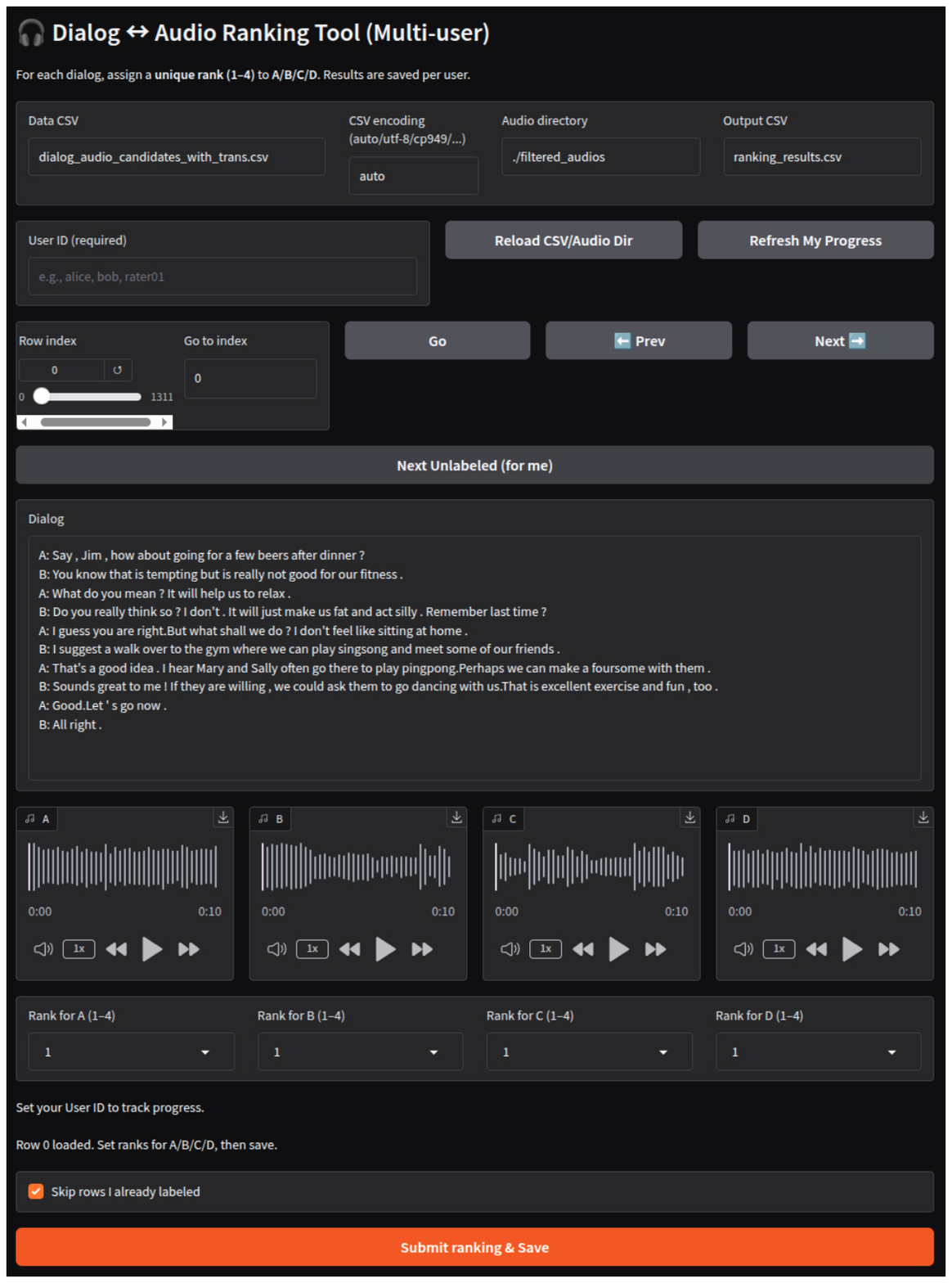} 
    \caption{Screenshot of the Gradio-based data collection interface. The tool displays the dialogue and allows annotators to rank four BGM candidates (Candidates A--D) before proceeding to the next turn.}
    \label{fig:gradio_interface}
\end{figure*}

\section{LLM Prompt for Dialogue Caption Generation}
\label{app:dialogue_caption_prompt}

To bridge the modality gap between multi-turn dialogues and concise music descriptions, we prompt an LLM to produce a single-sentence caption that encodes (i) the core topic of the dialogue, (ii) the implied conversational mood, (iii) salient keywords, and (iv) a brief background-music (BGM) suggestion. 

\begin{promptbox}[System Prompt: Dialogue-to-Caption for BGM Retrieval]
You are an expert annotator for background music selection. Given a two-speaker multi-turn dialogue, write exactly ONE natural sentence in English that covers: the core topic, 3--5 mood words, 3--5 concise keywords, and a short BGM suggestion (style/instrument/tempo/energy). Do not use lists, line breaks, quotes, or brackets. Keep the sentence between 20 and 35 words.
\end{promptbox}

\begin{promptbox}[User Prompt]
Read the dialogue below and produce exactly ONE sentence in English.

Rules:
- Include the core summary, and weave 3--5 mood words naturally (no brackets).
- Include 3--5 concise keywords, comma-separated, while keeping the sentence natural.
- Provide a brief BGM suggestion (style/instrument/tempo/energy) within the same sentence.
- No line breaks, no lists, and no quotes/brackets/braces.
- Output one sentence only.

Dialogue:
\{DIALOGUE\}
\end{promptbox}

\section{LLM Evaluation Prompt}~\label{appendix:prompt}
\label{sec:appendix_prompt}
For data construction, we also employ a structured prompt that elicits a continuous \([0.0, 10.0]\) suitability score for how well a candidate track functions as background music for a given dialogue. 

The prompt explicitly discourages central-tendency scoring, prioritizes technical interference checks (e.g., masking around speech frequencies), and enforces a strict JSON-only output format for reliable downstream parsing. It further specifies an internal (non-disclosed) scoring rubric that combines (A) acoustic technicality, (B) emotional/contextual fit, and (C) pacing/rhythm using a weighted formula, with additional constraints for prominent vocals/lyrics and poor technical compatibility.
The full system prompt used in our experiments is presented in the box below.

\begin{figure*}[ht!]
\begin{promptbox}[System Prompt for BGM Evaluation]
"You are a legendary Chief Audio Director and film music critic with 30+ years of experience in Hollywood. You are strict, detail-obsessed, and you do not compromise on quality.

Your task: judge how well the given candidate track works specifically as BACKGROUND MUSIC (BGM) for the dialogue context.

IMPORTANT RULES:
1) Reject central tendency: 'okay' is a lazy answer. Use the full 0–10 scale. Great matches deserve high scores; distracting or contradictory choices deserve low scores.
2) Technical interference first: before artistic taste, check for physical clashes.
   - Frequency masking risk: would the music likely intrude into speech intelligibility (roughly 1–4 kHz)?
   - Level balance risk: would the music likely overpower or compete with the dialogue?
   - If the track contains prominent vocals/lyrics, treat it as a Fatal Error for BGM under dialogue.
3) Narrative synchronization: the music must support the dialogue's emotional arc and setting.
4) Think step-by-step internally, but DO NOT reveal your reasoning.

Return ONLY a JSON object and nothing else.

Score how well this track fits as BACKGROUND MUSIC (BGM) for the given dialogue context.

OUTPUT REQUIREMENTS (STRICT):
- Return ONLY one JSON object with exactly one key: score.
- score must be a NUMBER in [0.0, 10.0] with EXACTLY ONE decimal place.
- Do NOT output an integer. Always include one decimal place.
- Use 0.1 granularity; avoid repeating the same score unless truly indistinguishable.

To avoid score clustering, compute the final score using this internal procedure (do NOT output the steps):
Step A — Acoustic Technicality (0.0–10.0, one decimal):
  Evaluate masking risk (busy midrange, sharp leads, dense percussion), perceived loudness competition, and how well it can sit under speech without needing aggressive ducking.
Step B — Emotional \& Contextual Fit (0.0–10.0, one decimal):
  Match mood, tension, warmth, setting, and emotional arc implied by the dialogue.
Step C — Pacing \& Rhythm (0.0–10.0, one decimal):
  Match energy/tempo to implied speech rate and scene pacing; penalize off-beat urgency/sleepiness.

Final score rule (compute numerically, then round to ONE decimal):
  base = 0.2*A + 0.5*B + 0.3*C
  If prominent vocals/lyrics are present (Fatal Error), then score = min(base, 2.0).
  If A < 3.0, then score must not exceed 3.0.
  subtle fit differences (e.g., slightly too busy -> -0.2; exceptionally well-undercored -> +0.2).
  Then clamp to [0.0, 10.0] and round to 1 decimal."
\end{promptbox}
\caption{The full system prompt used for LLM-based BGM evaluation. The prompt incorporates a weighted scoring mechanism ($0.2 \times \text{Technicality} + 0.5 \times \text{Fit} + 0.3 \times \text{Pacing}$) and explicit penalty rules for vocal interference.}
\label{fig:system_prompt}
\end{figure*}

\end{document}